\documentclass[letterpaper, 10 pt, conference]{support/IEEEconf} 
\usepackage{times}
\usepackage[pdftex]{graphicx}
\usepackage{amsmath,amssymb,amsopn,amstext,amsfonts}
\usepackage{pdfsync}
\usepackage{balance}
\usepackage{color}
\usepackage{algorithm}
\usepackage{algorithmic}
\usepackage{bm}
\usepackage{float}
\usepackage[linkcolor=black,citecolor=black,urlcolor=black,colorlinks=true]{hyperref}

\graphicspath{{figures/}}
\DeclareGraphicsExtensions{.pdf,.png,.jpg,.eps}
\IEEEoverridecommandlockouts

\title{\LARGE \bf Hybrid aerial-ground locomotion with a single passive wheel}

\author{Youming Qin, Yihang Li, Xu Wei, Fu Zhang 
\thanks{The authors are with the Department of Mechanical Engineering, University of Hong Kong. {\tt\small qym96@hku.hk}
}
}%

\begin{document}

\maketitle

\begin{abstract}
Exploiting contacts with environment structures provides extra force support to a UAV, often reducing the power consumption and hence extending the mission time. This paper investigates one such way to exploit flat surfaces in the environment by a novel aerial-ground hybrid locomotion.  Our design is a single passive wheel integrated at the UAV bottom, serving a minimal design to date. We present the principle and implementation of such a simple design as well as its control. Flight experiments are conducted to verify the feasibility and the power saving caused by the ground locomotion. Results show that our minimal design allows successful aerial-ground hybrid locomotion even with a less-controllable bi-copter UAV. The ground locomotion saves up to 77\% battery without much tuning effort. 
\end{abstract}

\section{Introduction}
With the ability to move in 3D spaces, multirotor UAVs have proved to be a successful platform for a variety of applications, such as aerial photographing, mapping, exploration, search and rescue \cite{khosiawan2016system, nex2014uav, baiocchi2013uav}. However, their limited onboard power dramatically decreases the flight time, especially at large payload \cite{8690666}. Even stationary hovering with a multirotor is energetically expensive as it requires constant motor actuation (hence power consumption) \cite{leishman2000principles}. Increasing the size of UAV propellers mitigate the problem, but its mobility in tight spaces will be seriously jeopardized \cite{9001153}.
\begin{figure}[h!]
\vspace{0.7cm}
    \begin{center}
        {\includegraphics[width=1.0\columnwidth]{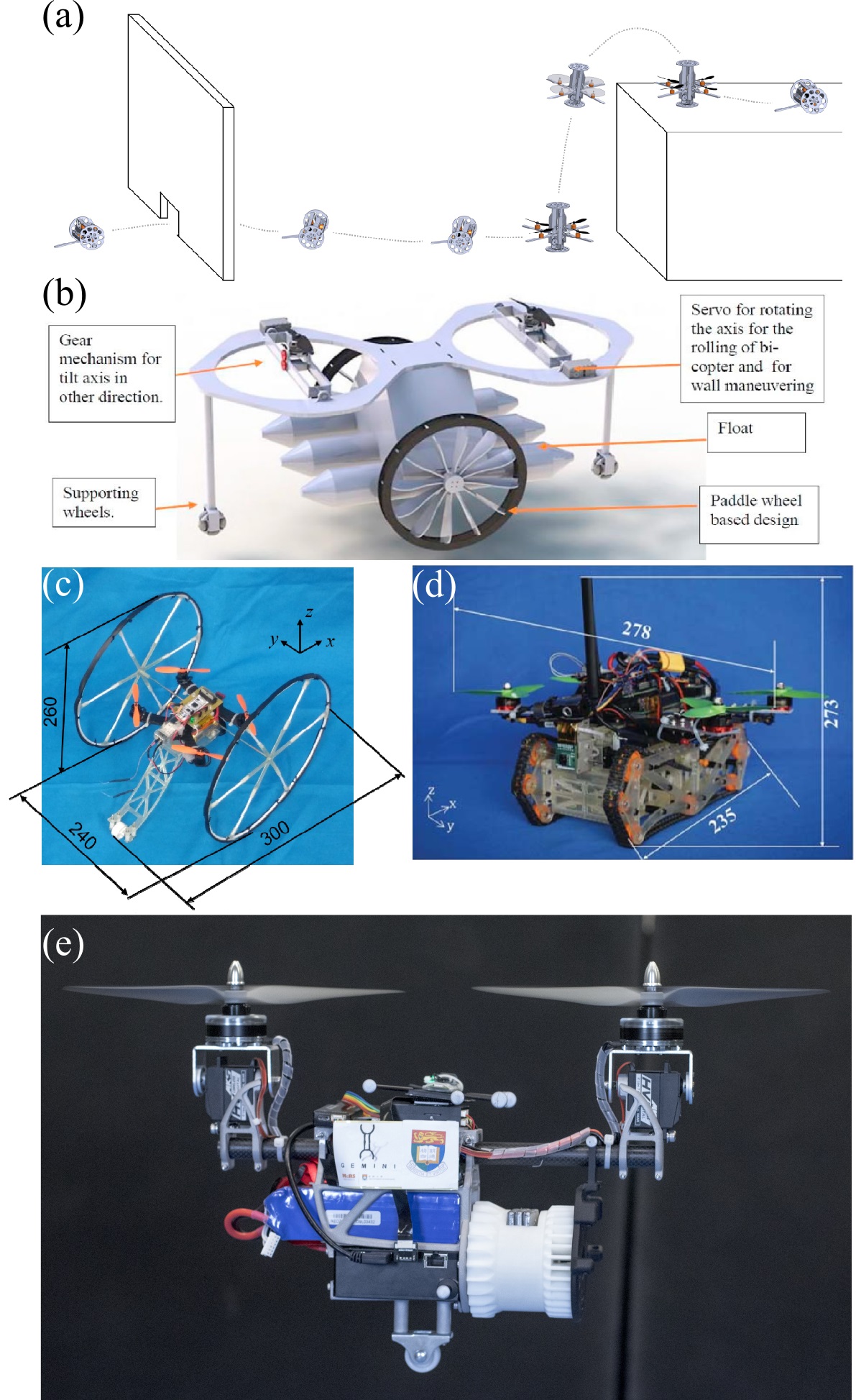}}
    \end{center}
    \vspace{-0.25cm}
    \caption{\label{fig:CommonHybridCompare}A review of common air-ground hybrid-motion vehicles: (a) Transformable hybrid ground-air vehicle \cite{8206402}, (b) Multi-terrain Multi-utility robot (MTMUR) \cite{adarsh2018multi}, (c) WAMORN (WAseda MOnitoring dRoNe) \cite{8016039}, (d) Air-ground amphibious agricultural information collection robot \cite{7734115}, (e) Our hybrid aerial-ground locomotion UAV with a single passive wheel.}
    \vspace{-0.75cm}
\end{figure}
Exploiting contacts with environmental structures could provide extra force support and reduce the required motor actuation (hence power). A widely researched approach is to perch a UAV on elevated locations (e.g., ceilings, walls, and others), mimicking the birds resting on tree branches or power lines \cite{kovac2016learning, graule2016perching, hang2019perching}. For example, Hang {\it et al.} \cite{hang2019perching} design a transformable landing gear with specially designed grippers which enable a multirotor to perch on different objects, poles, rooftops, branches, etc.. H. Zhang {\it et al.} \cite{8793936} propose a compliant bistable gripper that enables a micro quad-rotor Crazyflie 2.0 to perch on cylindrical objects. Other perching mechanisms such as dry-adhesive gecko-inspired grippers \cite{jiang2015perching, jiang2014modeling, thomas2016aggressive}, fiber-adhesive grippers \cite{daler2013perching} and dry-adhesive pads \cite{kalantari2015autonomous} have also been actively researched. Regardless of the exact implementation, a notable limitation of perching is that the UAV has to stay at the perching location. Traveling between those perching locations are still energetically expensive. 

Another approach is to obtain force support from the ground, leading to aerial-ground hybrid locomotion. Compared with perching, UAV moving on the ground cannot maintain a high vantage point but preserves the mobility, which is indeed necessary for tasks such as mobile mapping, exploration, search, and rescue. Although this concept is quite straightforward and long-existing in our lives, such as aircrafts landing gears for takeoff/landing, its formal application in aerial robots seems scarce. Most of the current work was motivated to improve the mobility of ground robots constrained by rough terrains in disaster relief \cite{adarsh2018multi,8016039, 7734115, 8206402} or out of entertainment or safety \cite{ParrotRollingSpider}. Specifically, Morton {\it et al.} \cite{8206402} presented a novel wheel-based locomotion (see Fig. \ref{fig:CommonHybridCompare}(a)). With a worm gear as actuator embedded in the middle, the robot can transform between driving mode to quad-rotors. Adarsh {\it et al.} \cite{adarsh2018multi} proposed an air-land-water vehicle concept with a wall maneuvering capability (see Fig. \ref{fig:CommonHybridCompare}(b)). The various functionality makes the design quite complicated, four wheels, two servo motors, and a chassis. This concept, unfortunately, has not come true to our best knowledge. Tanaka {\it et al.} \cite{8016039} developed a small mobile robot with a hybrid locomotion mechanism of wheels and multi-rotors, WAMORN (see Fig. \ref{fig:CommonHybridCompare}(c)). The design is very similar to the Parrot Rolling spider \cite{ParrotRollingSpider}, where two wheels are placed at the two sides of a quad-rotor.  A third wheel was also used for stabilizing the robot during running and allowing recovery from the flipping state. Wang {\it et al.} \cite{7734115} proposed a similar work of combining quad-rotors with chassis (see Fig. \ref{fig:CommonHybridCompare}(d)). This air-ground amphibious robot is aiming at operating in complex farmland terrains.

All the prior work mentioned above uses either at least two wheels or caterpillar chassis. Although they are able to move on rough terrains as expected, the driving mechanism adds up the significant weight to the system, leading to increased power consumption when moving in the air \cite{leishman2000principles}.

We take another view of the aerial-ground hybrid locomotion as a measure to improve UAV power efficiency while maintaining its mobility. Unlike the previous work, our approach is a single passive wheel installed at the bottom of the UAV, a minimal design among all others (see Fig. \ref{fig:CommonHybridCompare}(e)). Such a design simplifies the implementation and reduces the add-on weight: our preliminary implementation is 20 grams, adding merely 1\% to the UAV weight and minimizing the effect on flight power consumption. Moreover, it leads to a clear side view (i.e., a spinning LiDAR have all clear view on two sides, hence maximizing mapping efficiency).

Our contributions are the following: (1) we propose an aerial-ground hybrid locomotion with a single passive wheel, which is the minimal design so far; (2) we implement this design onto a bi-copter UAV \cite{9001153} and demonstrate its feasibility of enabling aerial-ground hybrid locomotion via real flight experiments; (3) we conduct extensive flight tests to validate the power saving enabled by the aerial-ground hybrid locomotion.

The rest of this paper is structured as follows: Section II elaborates the system design, core components onboard and implementation process; Section III explains the controller designs and the program that enables the air-ground transition; The experiment validation and power analysis will be presented in Section IV; in the end, the future work and conclusion are in Section V.

\section{System Design, and implementation}
In principle, the aerial-ground hybrid locomotion mechanism based on a single passive wheel is applicable to generic multi-rotor UAVs. In this paper, we take a bi-copter UAV known as Gemini \cite{9001153} as an illustrative example.  The design objective of the bi-copter platform is to provide a compact yet efficient UAV for carrying a 500 g 3D LiDAR and operating in tight indoor spaces. The propulsion system consists of two T-Motor MT4006 740KV brush-less motors, MKS HV1220 servo motors, T-Motor F35A 3-5S 32bit ESC, and APC10X45MR propellers. The previous design in \cite{9001153} used a 4S battery, limiting the maximum thrust produced by the two brush-less motors and hence the UAV maneuverability. To overcome this, we use a 5000 mAH 10C 5S battery with higher voltage in the current platform, resulting in a 300 g higher maximum thrust on each motor. The whole airframe and servo holders have also been rebuilt with 3D printed aluminum alloys to ensure sufficient rigidity. The total takeoff weight (with other components such as payload and markers as in Section. \ref{sec:experiment}) is 1950 g.

\begin{figure}[t]
\vspace{0cm}
    \begin{center}
        {\includegraphics[width=1\columnwidth]{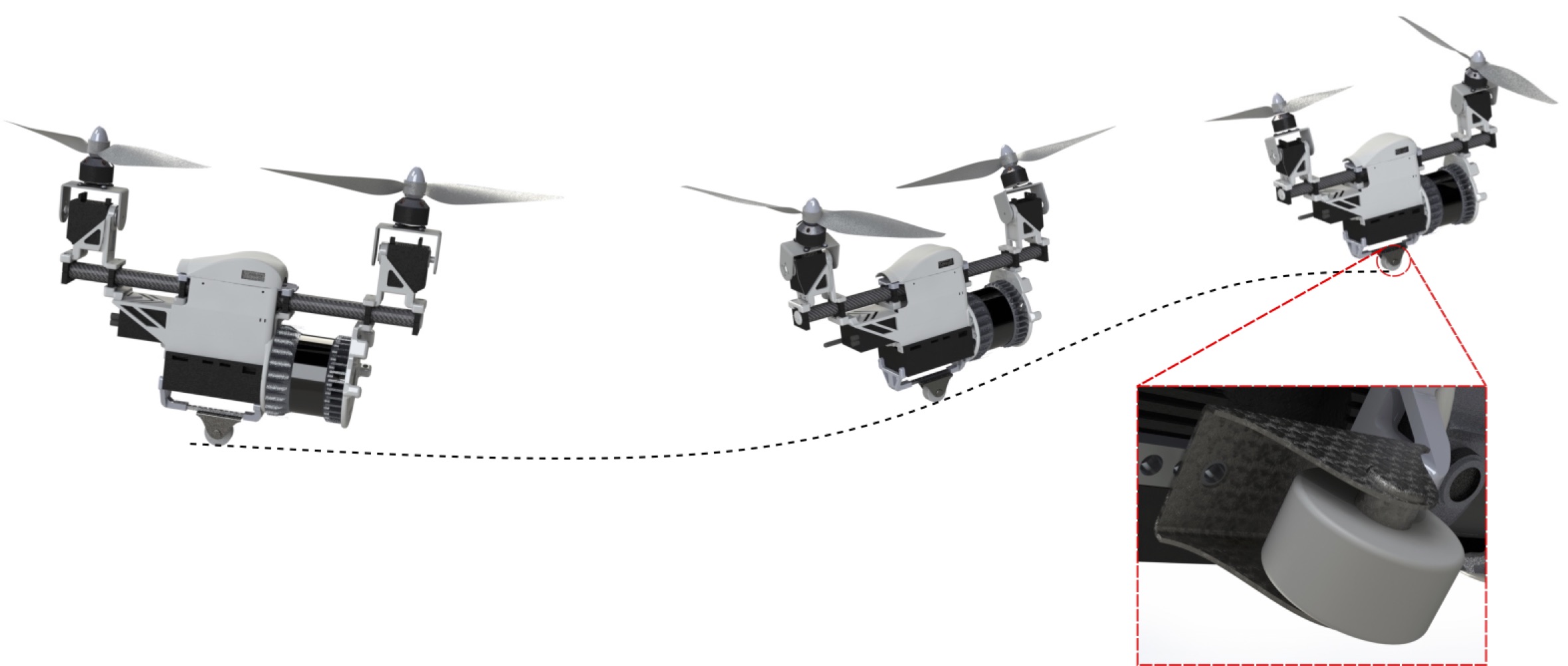}}
    \end{center}
    \vspace{-0.5cm}
    \caption{\label{fig:WorkingPrincipleSide} By changing the attitude and making use of the friction, the vehicle is able to move in any direction following an S-shaped curve.}
    \label{fig:wheel_design}
    \vspace{-1.5cm}
\end{figure}

In order to achieve aerial-ground hybrid locomotion with the minimal implementation, we picked an off the shelf nylon wheel of proper size. The wheel weighs 20 g and can roll along its shaft (see Fig. \ref{fig:wheel_design}). The wheel is attached to the UAV bottom right below the center of mass, minimizing the thrust difference of two brush-less motors when operating on the ground. Fig. \ref{fig:WorkingPrincipleSide} shows the working principle of the ground locomotion with a closeup look of the wheel. As shown in the figure, with the single wheel, the vehicle is able to move in any direction following an S-shaped curve by adjusting its attitude: the vehicle moves forward with the wheel rolling on the ground by pitching down and makes turns by changing the yaw directions. When making turns on the ground, the friction between the wheel and the ground will provide the centripetal force. Adjusting the UAV's attitude both in the air and on the ground is achieved by actuating the propeller thrust and servo angle (see \cite{9001153}). 

\section{Control}
\subsection{Dynamics Models}
\begin{figure}[h]
    \vspace{1cm}
    \begin{center}
        {\includegraphics[width=0.7\columnwidth]{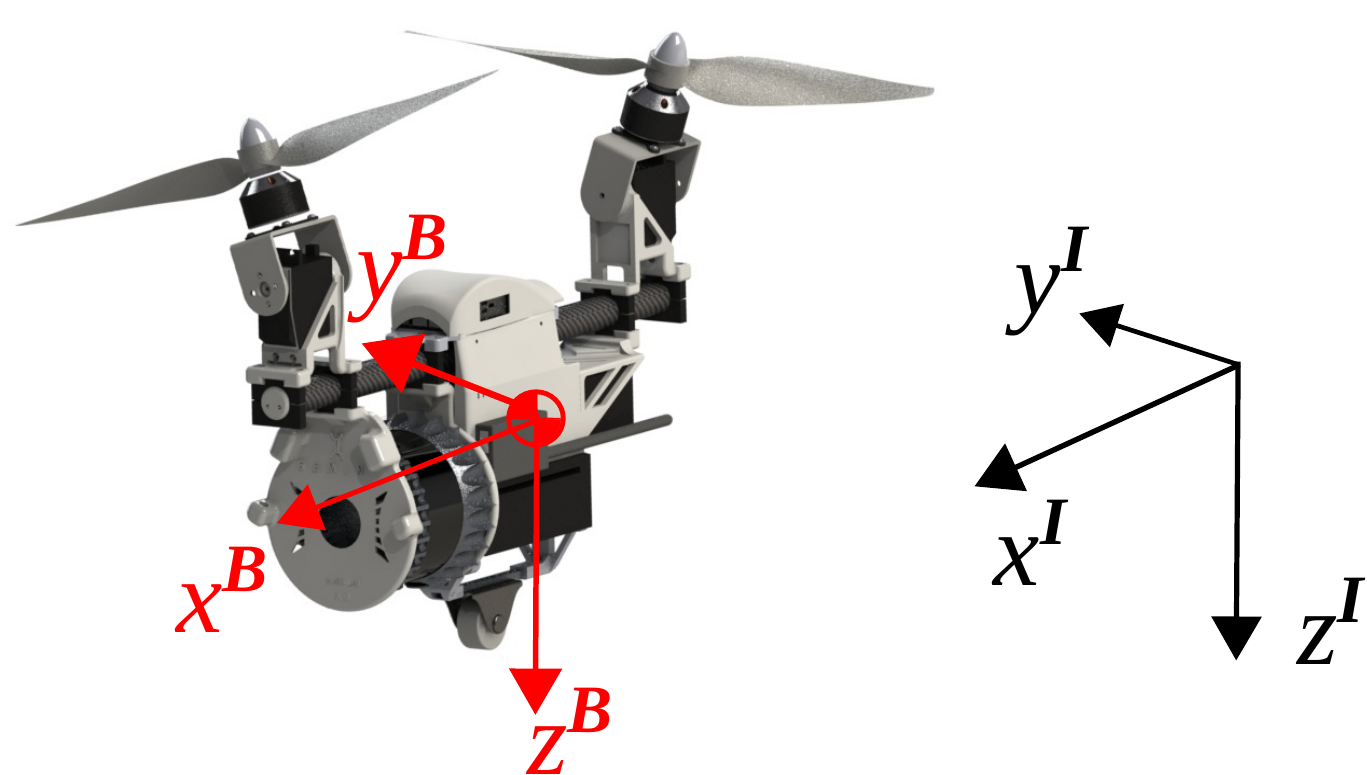}}
    \end{center}
    \vspace{0cm}
    \caption{\label{fig:frame}The definition of body frame ($x^\mathcal{B}y^\mathcal{B}z^\mathcal{B}$) and inertial frame ($x^\mathcal{I}y^\mathcal{I}z^\mathcal{I}$).}
    \vspace{0cm}
\end{figure}

\begin{figure}[h]
    \vspace{0cm}
    \begin{center}
        {\includegraphics[width=1\columnwidth]{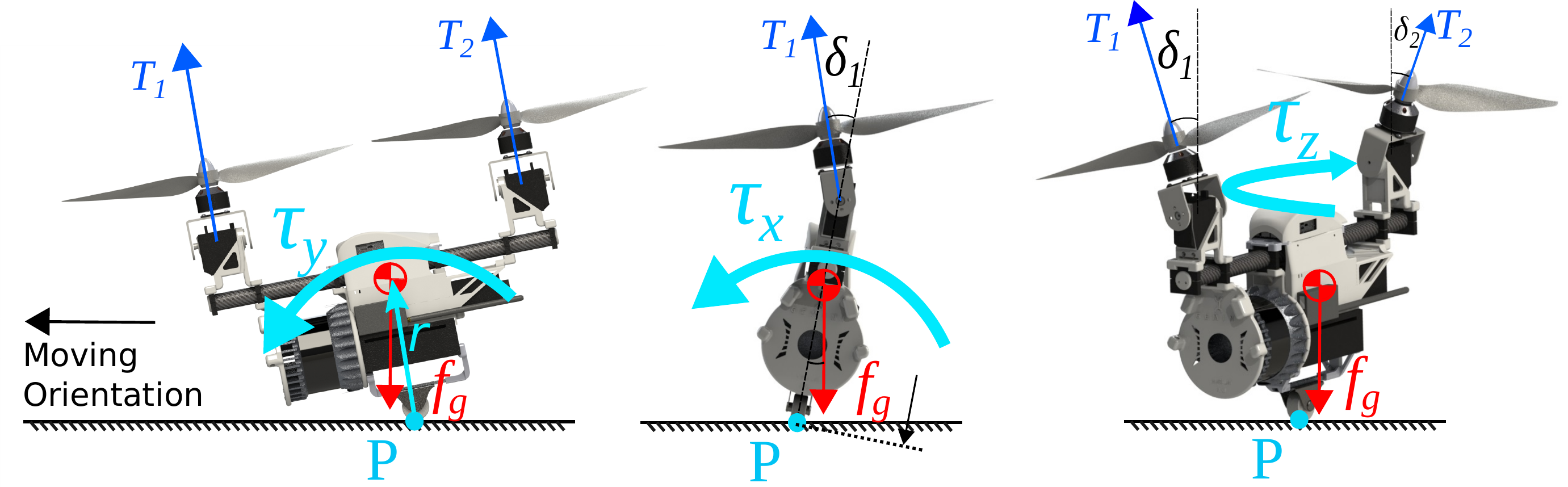}}
    \end{center}
    \vspace{-0.25cm}
    \caption{\label{fig:working_principle}The generation of the torque ($\tau_x$ $\tau_x$ $\tau_y$) in rolling mode.}
    \vspace{-0.25cm}
\end{figure}

Fig. \ref{fig:frame} shows the definition of body frame ($x^\mathcal{B}y^\mathcal{B}z^\mathcal{B}$) and N-E-D inertial frame ($x^\mathcal{I}y^\mathcal{I}z^\mathcal{I}$)\footnote{Throughout the text the superscript $^\mathcal{I}$ and $^\mathcal{B}$ will be used to denote the inertial and body frame, respectively.}. Since the vehicle dynamics in the air is identical to a bi-copter UAV, here we focus on the vehicle dynamics when rolling on the ground. The related notations are shown in Fig. \ref{fig:working_principle}. 

Assuming there is no slip between the wheel and the ground, the UAV rotates with respect to the contact point with the ground (the point P in Fig. \ref{fig:working_principle}). The dynamic model of this vehicle is as below:
\begin{subequations}\label{e:stand}
\begin{equation}\label{e:stand1}
m \dot{\bm v}_{CM}^\mathcal{I}=\bm f^\mathcal{I}
\end{equation}
\begin{equation}\label{e:stand2}
\bm{J}_{P}^\mathcal{B}\dot{\bm \omega}+{\widehat{\bm{\omega}}}{\bm{J}_{P}^\mathcal{B}}{{\bm{\omega}}}=\bm \tau^\mathcal{B}
\end{equation}
\end{subequations}
where the $m$ is the mass; the $\bm v_{CM}^\mathcal{I}$ denotes the velocity of the Center of Mass (CoM); the $\bm \omega$ stands for the angular rate represented in the body frame while the $\widehat{\bm \omega}$ is its skew-symmetric cross product matrix; the $\bm f^\mathcal{I}$ is the total force applied to the UAV including the gravity, thrust and other possible forces (e.g., frictions with the ground); the $\bm{J}_P^\mathcal{B}$ and $\bm \tau^\mathcal{B}=\left[\begin{matrix}\tau_x\quad\tau_y\quad\tau_z\end{matrix}\right]^T$ denote the inertia matrix and total torque with respect to the point P; Here denotes the torque generated by thrusts ($T_1$ and $T_2$ in Fig. \ref{fig:working_principle}) as $\bm \tau_t^\mathcal{B}$ whose detail can be found in our previous work \cite{9001153}. One of the main difference between the rolling dynamic and hovering dynamic is that the gravity will produce extra torque, as shown in Fig. \ref{fig:working_principle} and (\ref{e:rollingTau}):
\begin{equation}\label{e:rollingTau}
\begin{aligned}
\bm \tau^\mathcal{B} =\bm \tau_t^\mathcal{B}+\widehat{\bm r}^\mathcal{B} \bm f_g^\mathcal{B}
\end{aligned}
\end{equation}
where $\bm r^\mathcal{B}$ stands for the distance vector from point P to CoM that is a constant vector in body frame. For the translational motion on the rolling mode, the velocity of CoM (i.e., $\bm v_{CM}^\mathcal{I}$) should be:
\begin{equation}\label{e:rollingV}
\begin{aligned}
\bm v_{CM}^\mathcal{I}=\bm R\left(\bm v_P^\mathcal{B}+\widehat{\bm{\omega}}\bm r^\mathcal{B}\right)
\end{aligned}
\end{equation}
where $\bm R$ denotes the rotation matrix from the inertial frame to the body frame following the Z-Y-X Tait-Bryan order where $\eta$, $\theta$, and $\varphi$ stand for the yaw, pitch, and roll Euler angle, respectively, $\bm v_P^\mathcal{B}=\left[\begin{matrix}v_{Px}^\mathcal{B}\quad v_{Py}^\mathcal{B}\quad v_{Pz}^\mathcal{B}\end{matrix}\right]^T$ is the velocity of point P represented in body frame. It should be noted that no slipping between the wheel and the ground means $\bm v_P^\mathcal{B}=v_{Px}^\mathcal{B}\bm e_1$ where $\bm e_1=\left[1\quad 0\quad 0\right]^T$. Therefore, the derivation of CoM velocity is:
\begin{equation}\label{e:rollingVd}
\begin{aligned}
\dot{\bm v}_{CM}^\mathcal{I}=\bm R\left(\dot v_{Px}^\mathcal{B}\bm e_1+\widehat{\dot{\bm \omega}}\bm r^\mathcal{B}+\widehat{\bm \omega}v_{Px}^\mathcal{B}\bm e_1+\widehat{\bm{\omega}}^2\bm r^\mathcal{B}\right)
\end{aligned}
\end{equation}

Next, the total force consists of the friction force $\bm f_f$, support force $\bm f_N$, gravity force $\bm f_g$ and thrust force $\bm f_T$ as below
\begin{equation}\label{e:rollingF}
\begin{aligned}
\bm f^\mathcal{I}=\bm f_f^\mathcal{I}+\bm f_N^\mathcal{I}+\bm f_g^\mathcal{I}+\bm R\bm f_T^\mathcal{B}
\end{aligned}
\end{equation}

Putting (\ref{e:rollingTau}), (\ref{e:rollingVd}) and (\ref{e:rollingF}) back into (\ref{e:stand}), we can get the detailed dynamic model of rolling mode:
\begin{subequations}\label{e:full}
\begin{equation}\label{e:full1}
\begin{aligned}
m \bm R\left(\dot v_{Px}^\mathcal{B}\bm e_1+\widehat{\dot{\bm \omega}}\bm r^\mathcal{B}+\widehat{\bm \omega}v_{Px}^\mathcal{B}\bm e_1+\widehat{\bm{\omega}}^2\bm r^\mathcal{B}\right)= &\bm f_f^\mathcal{I}+\bm f_N^\mathcal{I}\\+&\bm f_g^\mathcal{I}+\bm R\bm f_T^\mathcal{B}
\end{aligned}
\end{equation}
\begin{equation}\label{e:full2}
\bm{J}_{P}^\mathcal{B}\dot{\bm \omega}+{\widehat{\bm{\omega}}}{\bm{J}_{P}^\mathcal{B}}{{\bm{\omega}}}=\bm \tau_t+\widehat{\bm r}^\mathcal{B} \bm f_g^\mathcal{B}
\end{equation}
\end{subequations}
\subsection{Controller design}
In order to implement automatic switching from aerial mode to rolling mode, additional control needs to be implemented. This section illustrate the detail of control process in each stage of the aerial-ground locomotion.
\subsubsection{Aerial mode}
\begin{figure}[ht]
\centering
\includegraphics[width=0.5\textwidth]{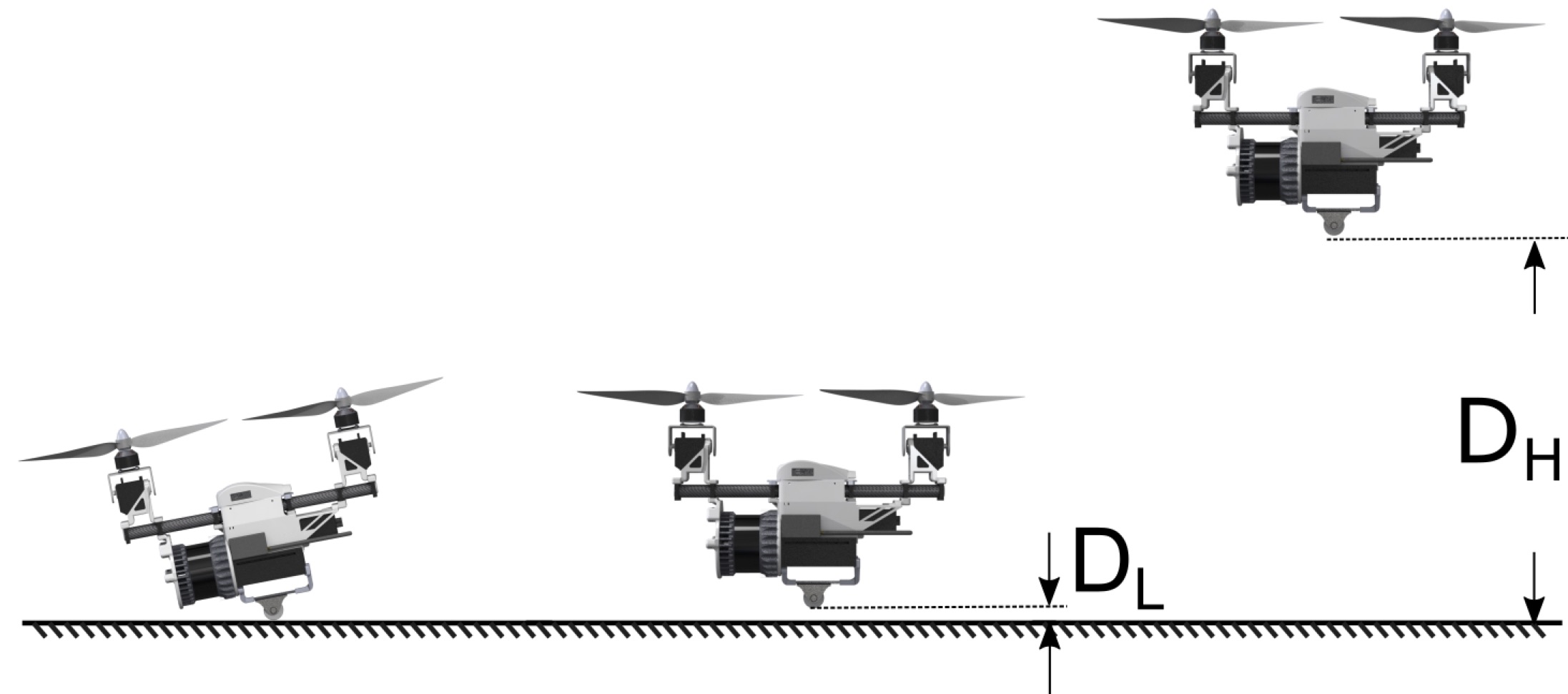}
\caption{Different phases of the hybrid aerial-ground locomotion: above $D_H$ is the aerial mode, below $D_L$ is the ground locomotion, and between them is the transition.}
\label{different heights}
\end{figure}

\begin{figure}[h]
\centering
\includegraphics[width=0.5\textwidth]{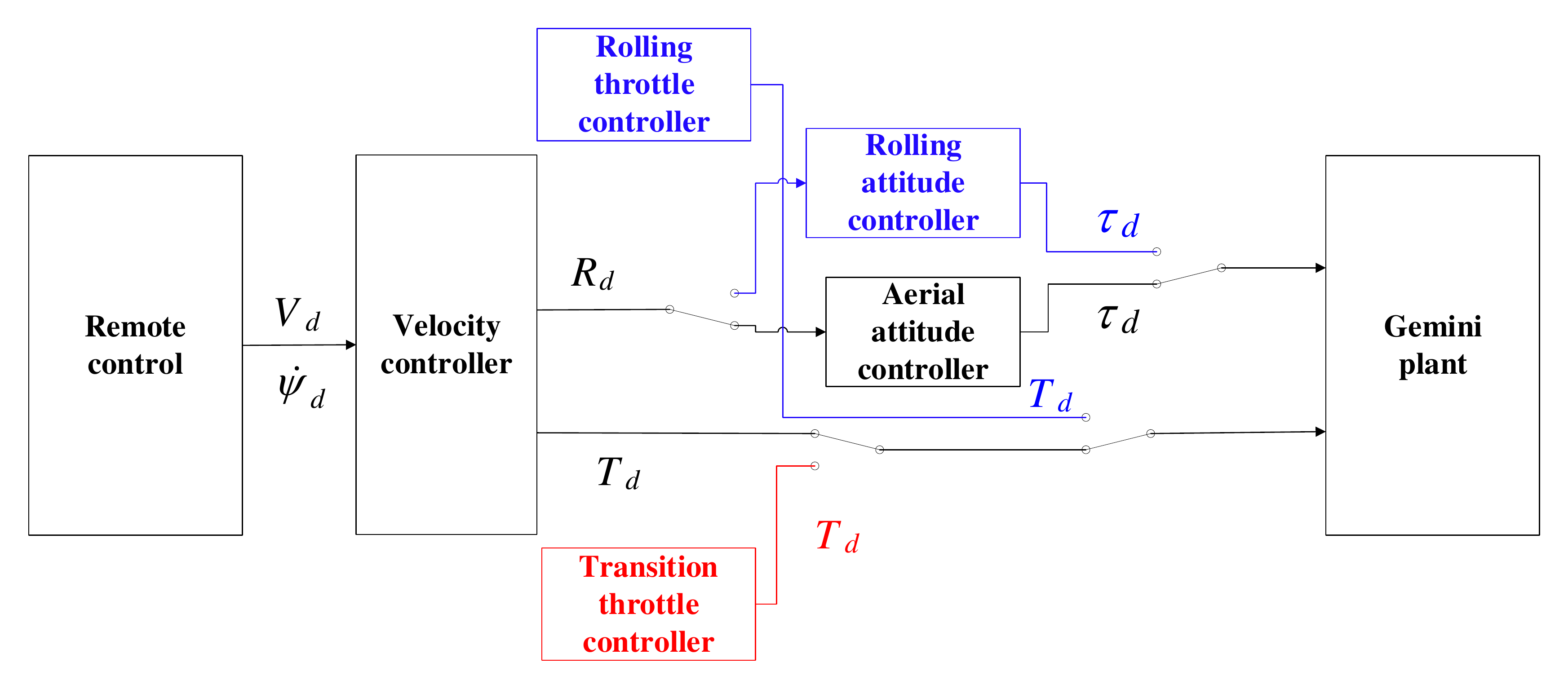}
\caption{Controller block diagrams: black blocks are used in normal aerial flight, red blocks are turned on when in transition mode and blue blocks are for control when rolling on the ground.}    
\label{Controlstructure}
\end{figure}

As shown in Fig. \ref{different heights}, the bi-copter is in aerial mode when it flies over a height $D_H$ or  where no ground effect is present. Its velocity and attitude are controlled through a cascaded controller as shown in Fig. \ref{Controlstructure}. The details about the velocity controller and attitude controller has been illustrated in our previous paper \cite{9001153}. In this process, pilot or mission planner commands the desired velocity $V_d$ and yaw rate $\dot\psi_d$.

\subsubsection{Transition mode}
When the pilot turns on transition switch, transition mode will be entered if the aircraft's height is lower than $D_H$. Once entered, the throttle command is controlled automatically in this mode. The aircraft approaches the ground by gravity when the throttle is lowered. To prevent a sudden falling, we set throttle command to gradually decreases. Because ground effect increases as the vehicle approaches the ground surface \cite{shi2019neural}, the throttle has to be decreased further to balance the additional lift caused by ground effect. The relation of throttle decreases with respect to time is described as:
\begin{equation}
T=T_h(1-Kt)
\end{equation}
where $T_h$ is hovering throttle near the ground, $K$ is the decreasing rate. Landing is automatically detected if the wheel touches the ground surface (i.e., the height of the wheel is below $D_L$ in \ref{different heights}) for a certain time. After this, the controller will switch to rolling mode.

\subsubsection{Rolling mode}
As mentioned previously, the dynamics of the rolling mode are considerably different from the aerial mode. The attitude can not be controlled using the same attitude control parameters as in aerial mode. Otherwise, the vehicle will go unstable and crashes within seconds. To overcome this, we use a new set of attitude controller parameters with the same structure (i.e., PID) as the aerial mode, which can be found in \cite{9001153}. The PID parameters are re-tuned based on our previous dynamics analysis. Due to additional torque caused by gravity in (\ref{e:full2}), the aircraft by itself is an unstable system like an inverted pendulum. As a result, the controller needs to have higher actuator values to compensate for the changes. Therefore we mainly tuned P and I and quickly got a good performance. The agility of rolling motion is determined by throttle command, and the ideal parameter is also obtained from the trial and error method.


\section{Experiment validation}\label{sec:experiment}

In order to validate the feasibility of our design in enabling aerial-ground hybrid locomotion, we conduct various flight experiments. Due to the space limit, we present the results in aerial-ground transition and ground locomotion and refer readers to our prior work \cite{9001153} for detailed aerial locomotion results. The experiments are conducted in an indoor environment equipped with OptiTrack motion capture systems providing the position feedback. An ultra wide-band module is used to transfer the position data to the onboard Pixhawk 4 mini flight controller. A full payload of 500 g is used to simulate the weight of a 16-line LiDAR (Ouster-16, weighing 380 g) and its connectors. The total takeoff weight of the whole UAV is 1950 g. Besides mobility tests, energy efficiency is also a core feature of hybrid locomotion. Therefore, we conduct a comparison study on power consumption in various conditions such as normal hovering, standing still on the ground, and rolling on the ground.

\subsection{Feasibility validation}

\subsubsection{Aerial-ground transition}
\quad
\par
\begin{figure}[t]
\vspace{+0.25cm}
    \begin{center}
        {\includegraphics[width=1.0\columnwidth]{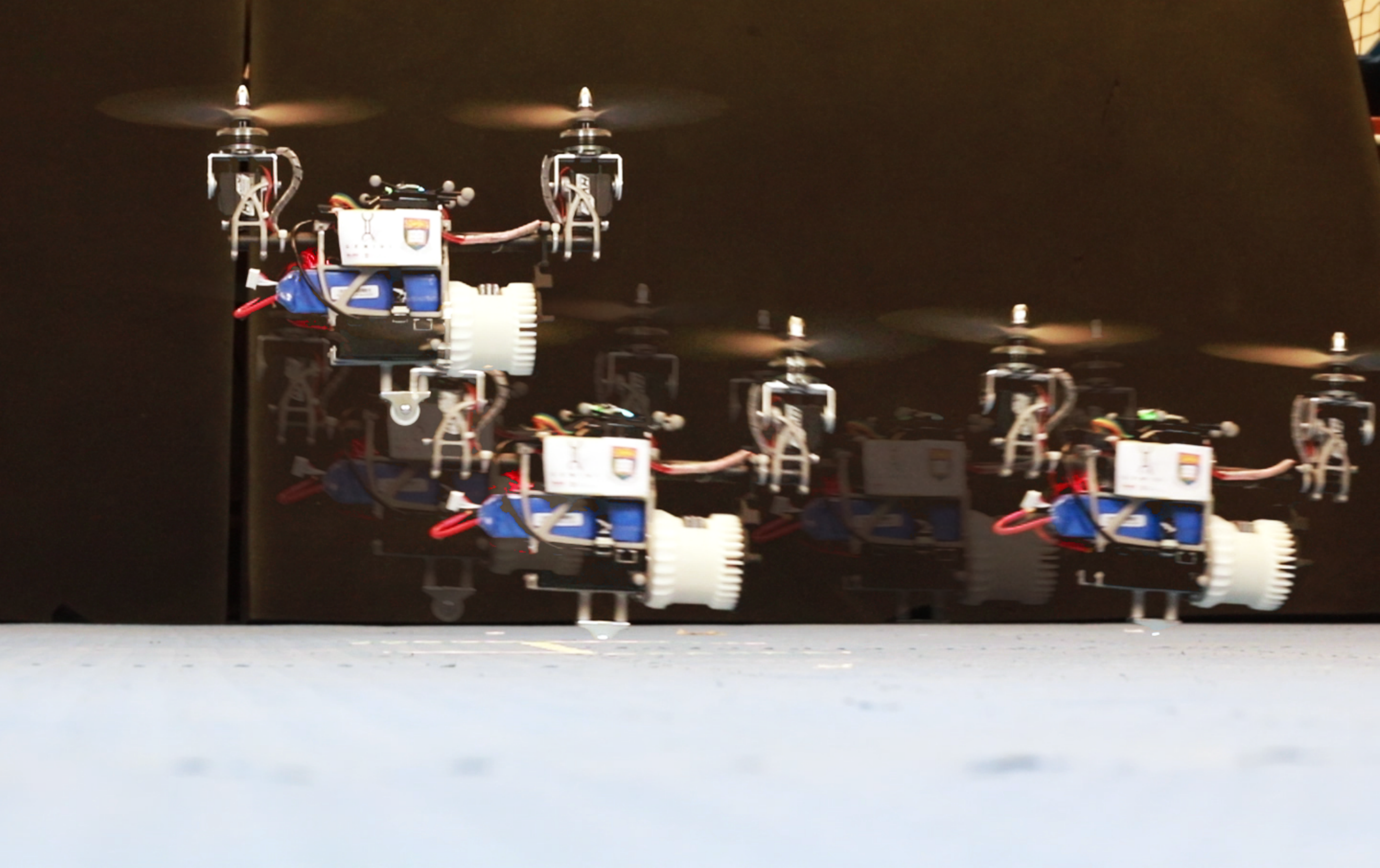}}
    \end{center}
    \vspace{-0.25cm}
    \caption{\label{fig:Transition}Aerial-ground transition process. }
    \vspace{-2cm}
\end{figure}
\begin{figure}[ht]
    \begin{center}
        {\includegraphics[width=0.9\columnwidth]{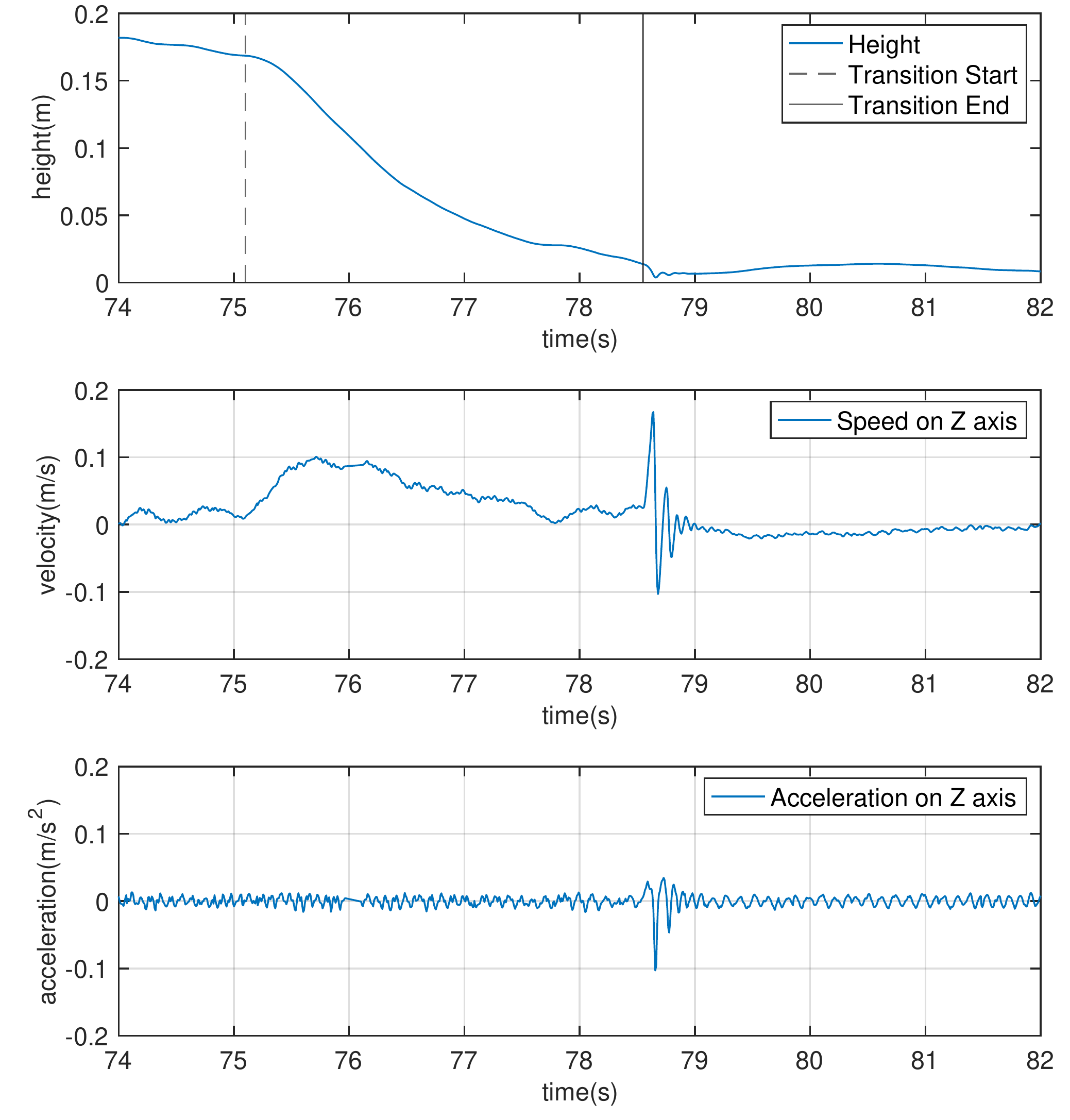}}
    \end{center}
    \vspace{-0.25cm}
    \caption{\label{fig:TransHeightVelocityAcceleration} The vertical position, velocity, and acceleration during aerial-ground transition. }
    \vspace{-0.25cm}
\end{figure}
\begin{figure}[ht]
    \begin{center}
        {\includegraphics[width=0.9\columnwidth]{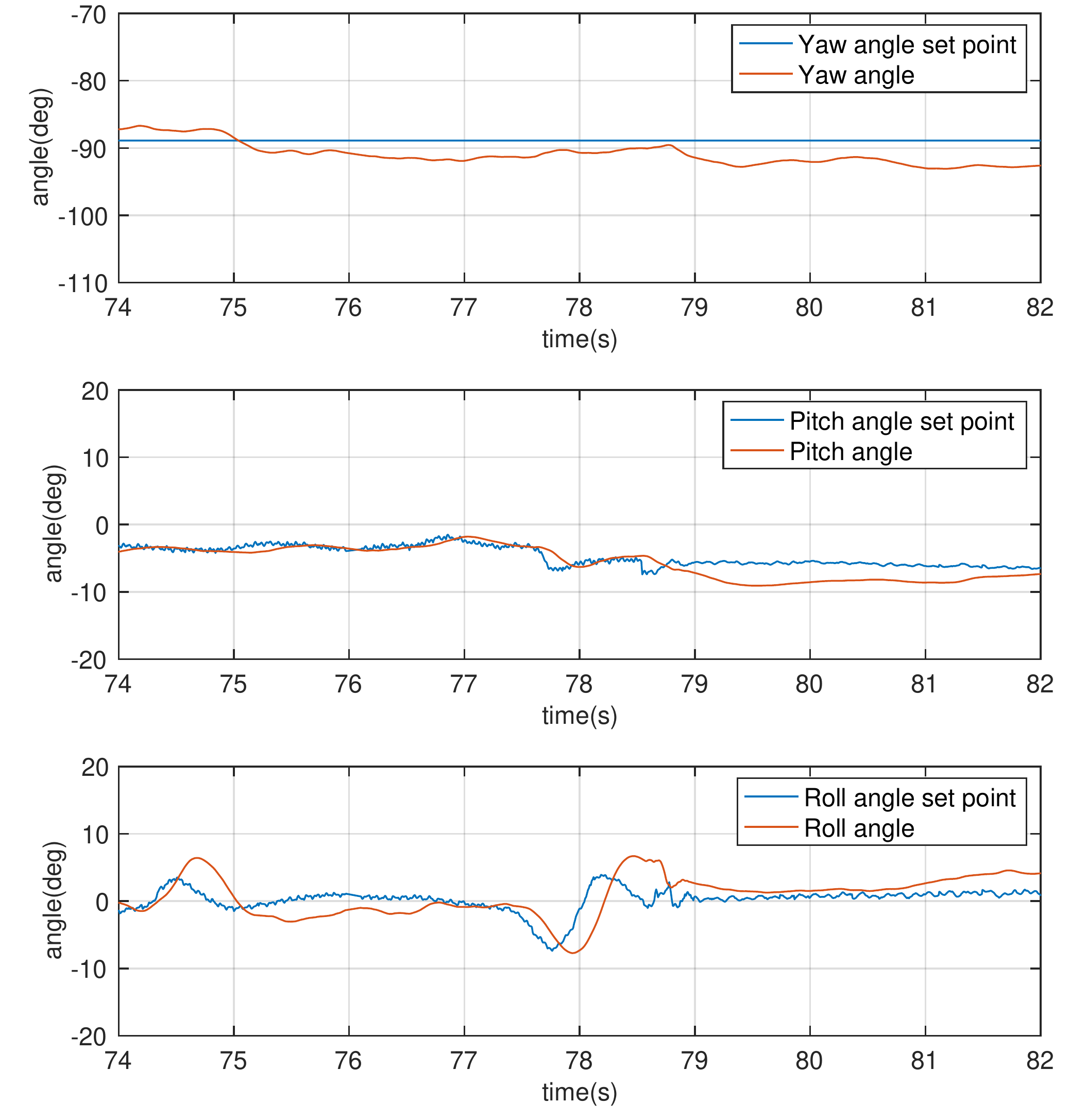}}
    \end{center}
    \vspace{-0.25cm}
    \caption{\label{fig:TransAngle} The attitude response during aerial-ground transition. }
    \vspace{-1.75cm}
\end{figure}

By using trial and error method during flight tests, we conclude that these following parameters has a high chance of soft touchdown: $D_H=0.2 m$ $D_L=0.01 m$ $K = 0.35 \%/s$ $\delta t = 0.15 s$. Fig. \ref{fig:Transition} shows the UAV poses at different times during the transition process overlaid together. The aircraft is initially hovering at the height of 15 cm above the ground and initiates a transition. It can be seen that the UAV successfully completes the transition and continues to move on the ground without interruption.  Fig. \ref{fig:TransHeightVelocityAcceleration} and Fig. \ref{fig:TransAngle} report the vertical translation and rotation during the landing process. After the landing command is sent out to the flight controller, the velocity on the Z direction increases to lower the height, and then decelerate until it reaches the ground surface to give the landing a soft touchdown. Due to the imprecision of the position feedback, the system received the signal of reaching the ground surface while it is actually a little bit above, throttle drops immediately, resulting in a spike oscillation afterward. The oscillation soon drops down and back to normal in half a second. The height data shows a little bounce back after the lowest point, but the whole transition process went on smoothly, and the bounce oscillation is barely observable by eyes. The attitude control is close to the idea. Even though the touchdown shocks the roll angle, it rapidly recovered to normal within half a second. The success of this experiment proves the feasibility of the concept of hybrid aerial-ground locomotion with a single passive wheel on a bi-copter.

\subsubsection{Ground locomotion}
\quad
\par
\begin{figure}[h]
    \begin{center}
        {\includegraphics[width=0.9\columnwidth]{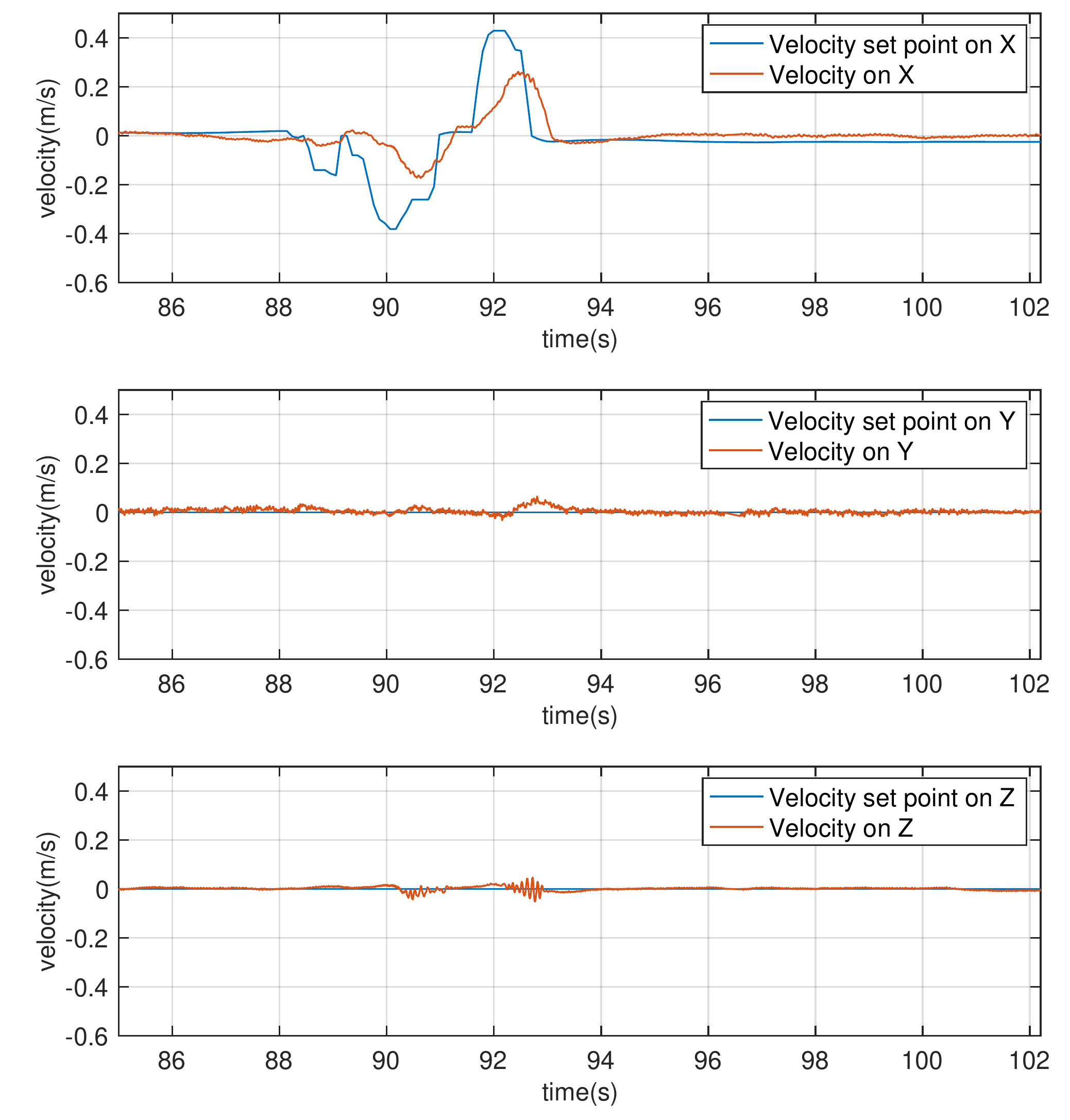}}
    \end{center}
    \vspace{-0.25cm}
    \caption{\label{fig:RollingVelocity} Velocity response when rolling on ground. }
    \vspace{-0.25cm}
\end{figure}
\begin{figure}[h]
    \begin{center}
        {\includegraphics[width=0.9\columnwidth]{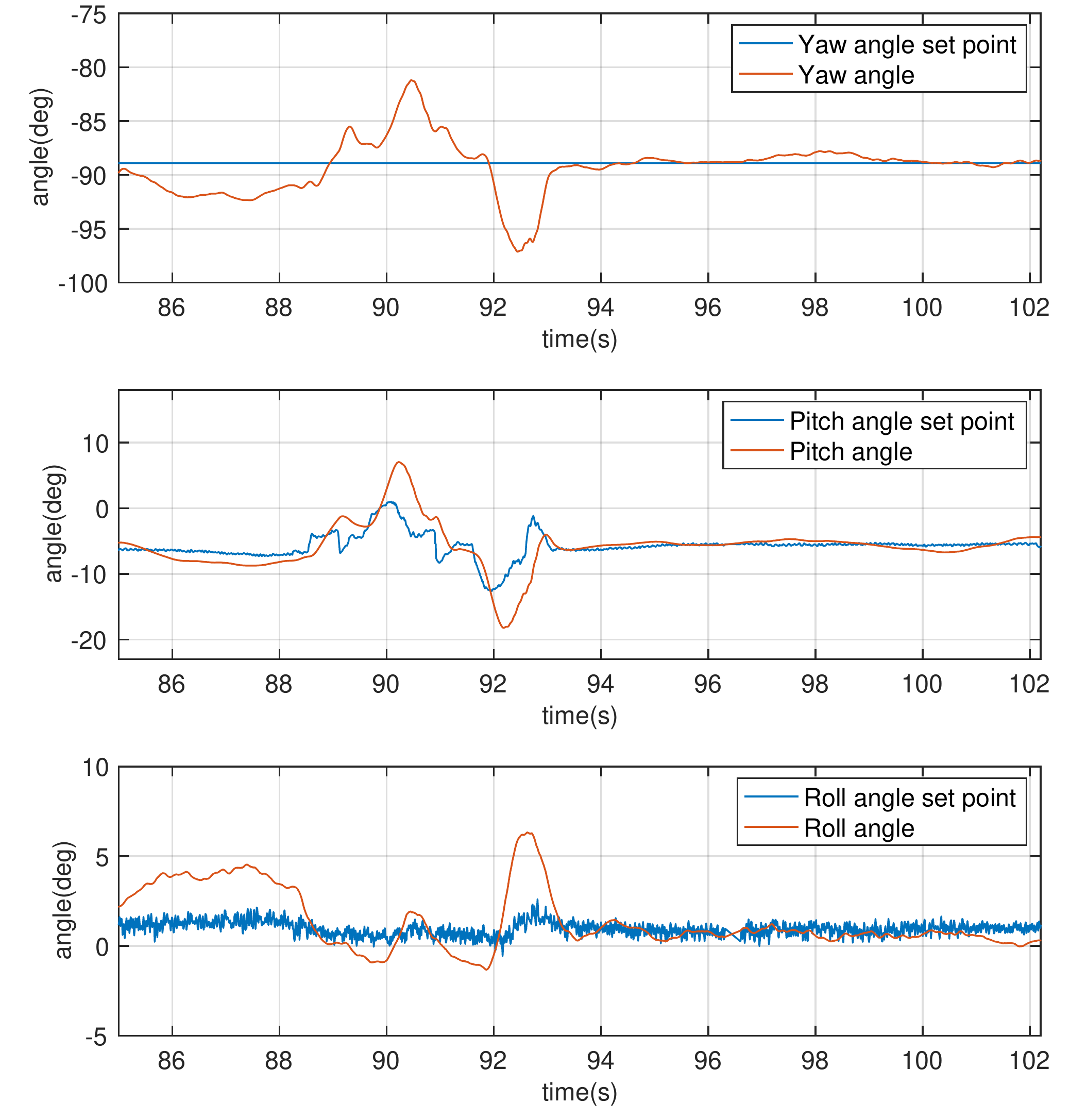}}
    \end{center}
    \vspace{-0.25cm}
    \caption{\label{fig:RollingAngle} Attitude response when rolling on ground. }
    \vspace{-1cm}
\end{figure}

Fig. \ref{fig:RollingVelocity} and Fig. \ref{fig:RollingAngle} shows the velocity and attitude response when the aircraft is rolling on the ground back and forward. Velocity in the body X direction is commanded while the rest are set to zero. As expected, the velocity profiles in all three directions are tracked well. Inspecting the attitude responses, we found what the pitch command is varying to achieve the translation along that direction while the rests are around zero; all are within our expectation. The attitude responses suffer from some fluctuations, especially in yaw and roll, but overall remain stable. These fluctuations are due to the friction between the wheel and ground, which is not seriously compensated in our attitude controller. Improving the attitude controller performance by carefully modeling and exploiting these contact forces will be interesting future work.

\begin{figure}[ht]
    \begin{center}
        {\includegraphics[width=1.0\columnwidth]{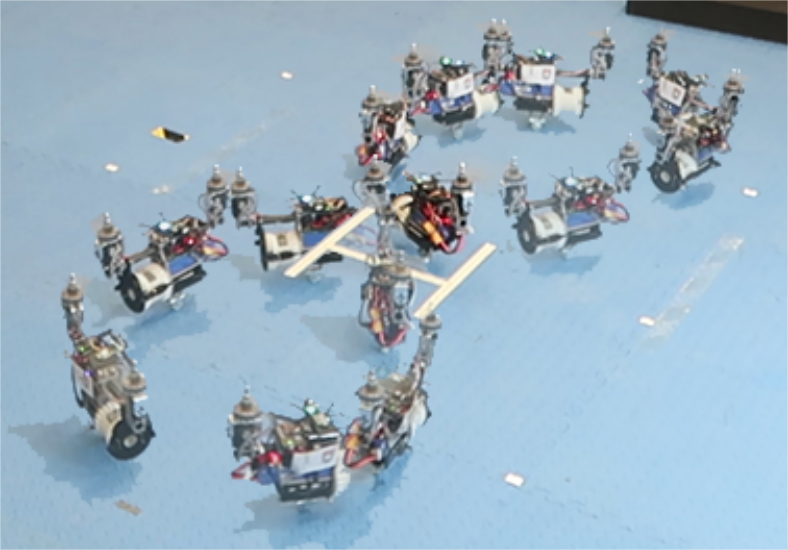}}
    \end{center}
    \vspace{-0.25cm}
    \caption{\label{fig:8} Rolling on the ground following a pattern ``8". }
\end{figure}

Besides following a straight line, we also command the vehicle to follow an ``8" figure on the ground. Due to space limitations, the data is not presented. Instead, the whole process is overlaid and shown in Fig. \ref{fig:8}. Readers may also refer to the submitted video.  

\subsection{Efficiency demonstration}
\subsubsection{Power consumption comparison}
\quad
\par
\begin{figure}[ht]
    \begin{center}
        {\includegraphics[width=0.7\columnwidth]{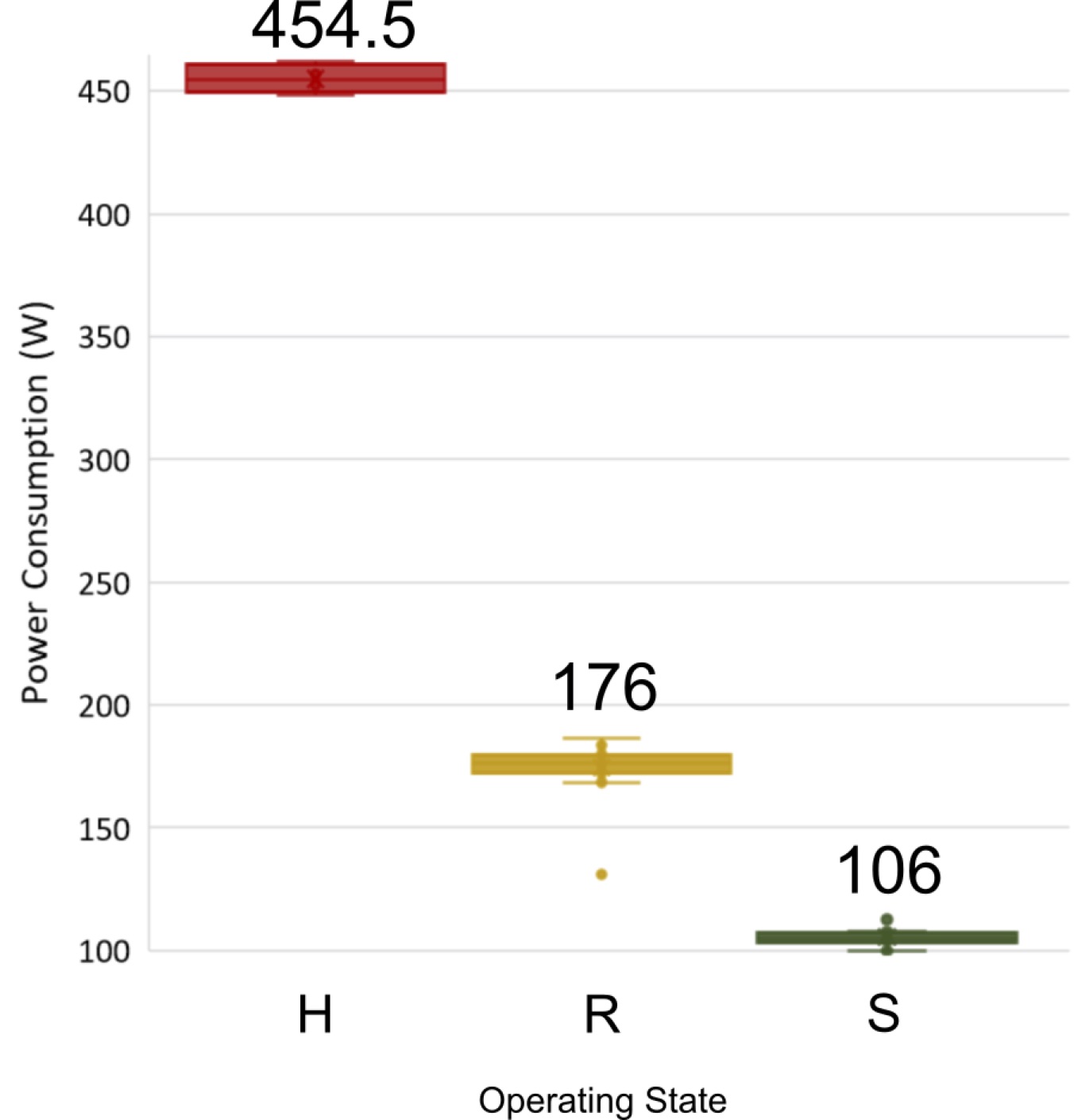}}
    \end{center}
    \vspace{-0.25cm}
    \caption{\label{fig:PowerConsumption}The power consumption comparison among different states: hovering state (H); rolling on the ground state (R); sitting on the wheel state (S). The numbers above each bar are the mean power consumption of the state over 10 seconds. }
    \vspace{-0.25cm}
\end{figure}

Empirical power consumption comparison among different flight phases are reported in Fig. \ref{fig:PowerConsumption}. The power consumption data are gathered when the aircraft is hovering high above the ground surface with no ground effect (H state), rolling on the ground surface (R state), and sitting still on the ground surface, respectively (S state). It can be seen from the figure that sitting on the ground consumes the least power. Rolling on the ground costs a bit more energy than sitting as the aircraft has to overcome the friction. In conclusion, when comparing to the H state, R state, and S states save 61, and 77\% power, respectively. This proves the considerable energy saving with the ground locomotion.

\section{Conclusion and Future work}
This paper proposes novel aerial-ground hybrid locomotion with a single passive wheel. The design, implementation, control, and flight experiment are presented. Unlike previous works that sacrifice weight for heavy chassis to achieve hybrid locomotion, our approach achieves great power efficiency advantages of the ground locomotion with the minimal addition of a single wheel that weighs only 1\% of the total takeoff weight. Flight experiments validated the feasibility and efficiency of our proposal. 

A drawback of the single wheel design is that the system is inherently unstable and cannot land on the ground. So an extra landing station or more sophisticated landing strategies or mechanisms are required. The inherently unstable system dynamics may also lower the vehicle stability when moving on rough terrains. Even with flat ground, the current control strategy involves complicated switching strategies and relies on many human tuning efforts. Future work will focus on systematic modeling and control of the vehicle dynamics on the ground, especially taking friction into consideration, to enable smooth, agile, and stable ground locomotion and aerial-ground transition.   

\bibliography{root} 

\end{document}